# Uniqueness Domains in the Workspace of Parallel Manipulators

Philippe WENGER, Damien CHABLAT

Laboratoire d'Automatique de Nantes URA CNRS 823 (École Centrale de Nantes / Université de Nantes)
1, rue de la Noë, BP 92101, 44321 Nantes Cedex 3 France
email : Philippe.Wenger@lan.ec-nantes.fr

## ABSTRACT

This work investigates new kinematic features of parallel manipulators. It is well known that parallel manipulators admit generally several direct kinematic solutions for a given set of input joint values. The aim of this paper is to characterize the uniqueness domains in the workspace of parallel manipulators, as well as their image in the joint space. The study focuses on the most usual case of parallel manipulators with only one inverse kinematic solution. The notion of *aspect* introduced for serial manipulators in **[Borrel 86]** is redefined for such parallel manipulators. Then, it is shown that it is possible to link several solutions to the forward kinematic problem without meeting a singularity, thus meaning that the aspects are not uniqueness domains. An additional set of surfaces, namely the *characteristic surfaces*, are characterized which divide the workspace into *basic regions* and yield new uniqueness domains. This study is illustrated all along the paper with a 3-RPR planar parallel manipulator. An octree model of spaces is used to compute the joint space, the workspace and all other newly defined sets.

**Key-words** : Parallel Manipulator, Workspace, Singularity, Aspect, Uniqueness domains, Octree.

## 1. INTRODUCTION

A well known feature of parallel manipulators is the existence of multiple solutions to the direct kinematic problem. That is, the mobile platform can admit several positions and orientations (or configurations) in the workspace for one given set of input joint values **[Merlet 90]**. The dual problem arises in serial manipulators, where several input joint values correspond to one given configuration of the end-effector. To cope with the existence of multiple inverse kinematic solutions in serial manipulators, the notion of *aspects* was introduced **[Borrel 86]**. The aspects were defined as the maximal singularity-free domains in the joint space. For usual industrial serial manipulators, the aspects were found to be the maximal sets in the joint space where there is only one inverse kinematic solution. Many other serial manipulators, referred to as *cuspidal* manipulators, were shown to be able to change solution without passing through a singularity, thus meaning that there is more than one inverse kinematic solution in one aspect. New uniqueness domains have been characterized for cuspidal manipulators **[Wenger 92]**, **[El Omri 96]**. It is also of interest to be able to characterize the uniqueness domains for parallel manipulators, in order to separate and to identify, in the workspace, the different solutions to the direct kinematic problem. To the authors knowledge, the only work concerned with this issue is that of **[Chételat 96]**, which proposes a generalization of the implicit function theorem. Unfortunately, the hypothesis of convexity required by this new theorem is still too restrictive. This paper is organized as follows. Section 2 describes the planar 3-RPR parallel manipulator which will be used all along this paper to illustrate the new theoretical results. Section 3 restates the notion of *aspect* for parallel manipulators. New surfaces, the *characteristic surfaces*, are defined in section 4, which, together with the singular surfaces, further divide the aspects into smaller regions, called *basic regions*. Finally, the uniqueness domains are defined in section 5. The workspace, the aspects, the characteristic and singular surfaces, and the uniqueness domains are calculated for the planar 3-RPR parallel manipulator using octrees. The images in the joint space of the uniqueness domains are also calculated. It is shown that the joint space is composed of several subspaces with different numbers of direct kinematic solutions.

## 2. PRELIMINARIES

### 2.1 PARALLEL MANIPULATOR STUDIED

This work deals with those parallel manipulators which have only one inverse kinematic solution. In addition, the passive joints will be always assumed unlimited in this study. For more legibility, a planar manipulator will be used as illustrative example all along this paper. This is a planar 3-DOF manipulator, with 3 parallel RPR chains (Figure 1). The input joint variables are the three prismatic actuated joints. The output variables are the positions and orientation of the platform in the plane. This manipulator has been frequently studied, in particular by **[Merlet 90]**, **[Gosselin 91]** and **[Innocenti 92]**.

The kinematic equations of this manipulator are **[Gosselin 91]** :

$$\rho_1^2 = x^2 + y^2 \qquad \textit{(1)}$$

$$\rho_2^2 = \left(x + l_2\, cos(\phi) - c_2\right)^2 + \left(y + l_2\, sin(\phi)\right)^2 \qquad \textit{(2)}$$

$$\rho_3^2 = \left(x + l_3\, cos(\phi + \theta) - c_3\right)^2 + \left(y + l_3\, sin(\phi + \theta) - d_3\right)^2 \qquad (3)$$

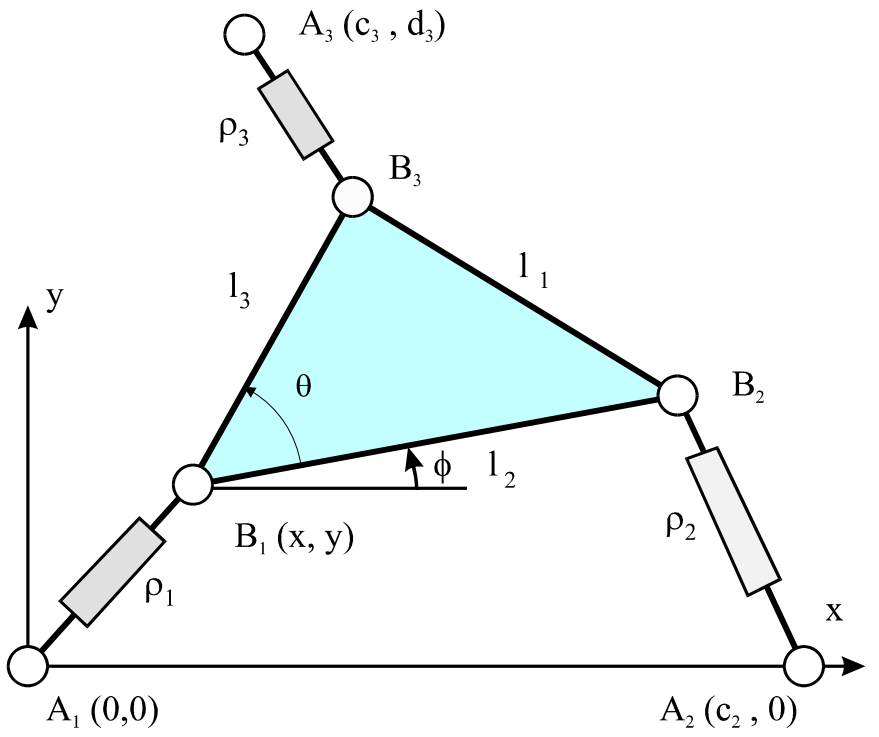

*Figure 1 : 3-RPR planar manipulator*

The dimensions of the platform are the same as in **[Merlet 90]** and in **[Innocenti 92]** :

- $A_1$= (0.0, 0.0) $B_1B_2$= 17.04
- $A_2$= (15.91, 0.0) $B_2B_3$= 16.54
- $A_3$= (0.0, 10.0) $B_3B_1$= 20.84

The limits of the prismatic actuated joints are those chosen in **[Innocenti 92]** :

$$10.0 \leq \rho_i \leq 32.0$$

The passive revolute joints are assumed unlimited.

## 2.2 OCTREE MODEL

The octree is a hierarchical data structure based on a recursive subdivision of space **[Meagher 81]**. It is particularly useful for representing complex 3-D shapes, and is suitable for Boolean operations like union, difference and intersection. Since the octree structure has an implicit adjacency graph, arcwise-connectivity analyses can be naturally achieved. The octree model of a space S leads to a representation of S with cubes of various sizes. Basically, the smallest cubes lie near the boundary of the shape and their size determines the accuracy of the octree representation. Octrees have been used in several robotic applications **[Faverjon 84]**, **[Garcia 89]**, **[El Omri 93]**. In this work, the octree models are calculated using discretization and enrichment techniques as described in **[Chablat 96]**.

The octree models of the reachable joint space (in the space $\rho_1$, $\rho_2$, $\rho_3$) and of the workspace (in the space $x$, $y$ et $\phi$) of the 3-RPR parallel manipulator are shown in figures 2 and 3 (the workspace and the reachable joint space are defined in section 3.1). The reachable joint space is not a complete parallelepiped, since not any joint vector can lead to an assembly configuration of the manipulator.

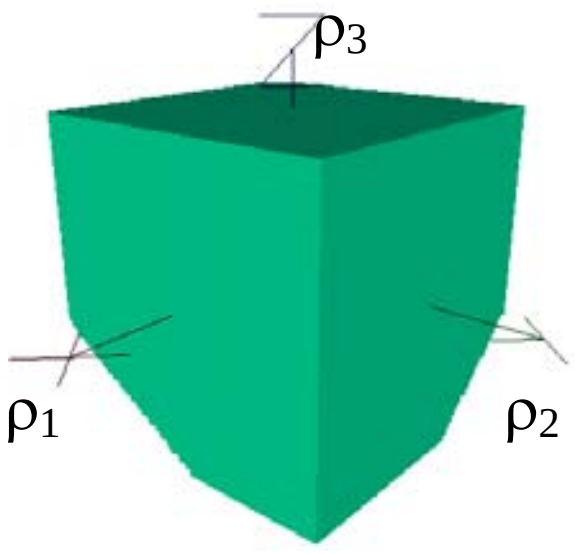

*Figure 2 : Octree model of the joint space*

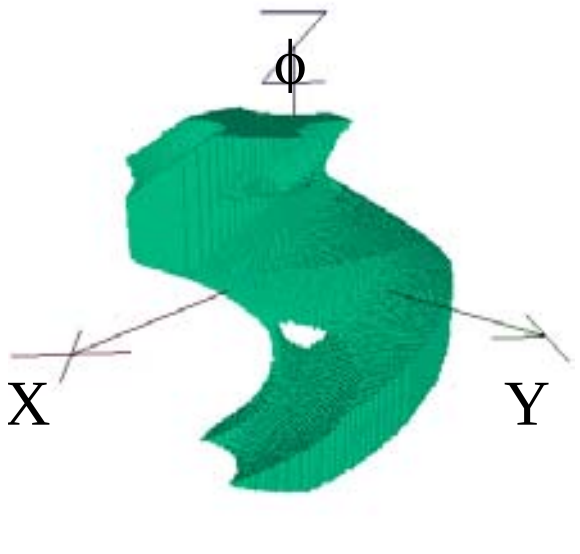

*Figure 3 : Octree model of the workspace*

## 2.3 SINGULARITIES

The vector of input variables and the vector of output variables for a n-DOF parallel manipulator are related through a system of non linear algebraic equations which can be written as :

$$F(q, X) = 0 \qquad (4)$$

where *0* means here the n-dimensional zero vector.

Differentiating (4) with respect to time leads to the velocity model :

$$A\dot{X} + B\dot{q} = 0 \qquad (5)$$

where $A$ et $B$ are $n \times n$ Jacobian matrices. These matrices are functions of $q$ and $X$ :

$$A = \frac{\partial F}{\partial X} \qquad B = \frac{\partial F}{\partial q} \qquad (6)$$

These matrices are useful for the determination of the singular configurations **[Sefrioui 92]**.

### 2.3.1 Type-1 singularities

These singularities occur when $det(B) = 0$.

For the planar manipulator, this condition can be satisfied only when $\rho_1 = 0$ or $\rho_2 = 0$ or $\rho_3 = 0$.

In practise, the type-1 singularities are attained when one of the actuated prismatic joints reaches its limit [**Gosselin 90**]. The corresponding configurations are located at the boundary of the workspace.

For parallel manipulators which may have more than one inverse kinematic solution, type-1 singularities are configurations where two solutions to the inverse kinematic problem meet. By hypothesis, type-1 singularities will be always associated with joint limits in this paper.

### 2.3.2 Type-2 singularities

They occur when $det(A) = 0$. Unlike the preceding ones, such singular configurations occur inside the workspace. They correspond to configurations for which two branches of the direct kinematic problem meet. They are particularly undesirable since the manipulator cannot be steadily controlled in such configuration where the manipulator stiffness vanishes in some direction.

For the planar manipulator, such configurations are reached whenever the axes of the three prismatic joints intersect (possibly at infinity). In such configurations, the manipulator cannot resist a wrench applied at the intersecting point (Figure 4).

The resulting singular surface is built and modelled using octrees (Figure 5). The equation of *Det(A)* can be put in an explicit form $y = s(x,\phi)$, only two variables need to be swept in the octree enrichment process **[Chablat 96]**.

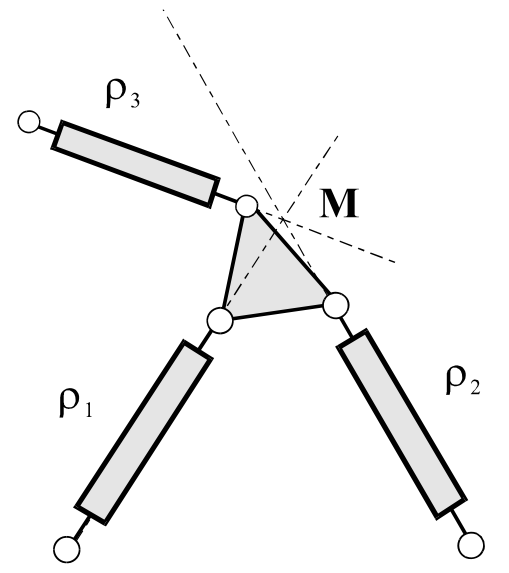

*Figure 4 : Type-2 singular configurations*

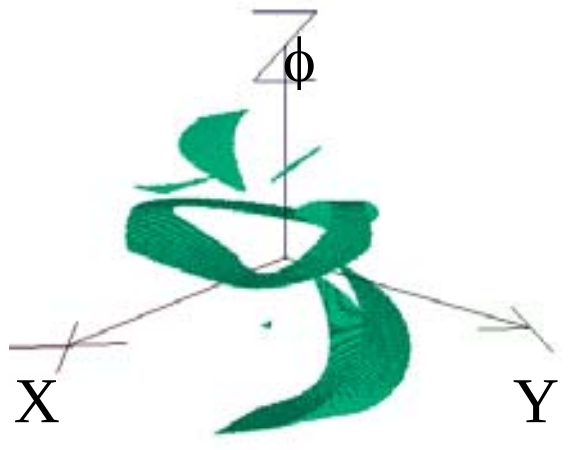

*Figure 5 : Octree model of the set of type-2 singularities in the workspace*

## 3. NOTION OF ASPECT FOR PARALLEL MANIPULATORS

The notion of aspect was introduced by **[Borrel 86]** to cope with the existence of multiple inverse kinematic solutions in serial manipulators. An equivalent definition was used in **[Khalil 96]** for a special case of parallel manipulators, but no formal, more general definition has been set. The aspects are redefined formally in this section.

### 3.1 DEFINITION

Let $OS_m$ and $JS_n$ denote the operational space and the joint space, respectively. $OS_m$ is the space of configurations of the moving platform and $JS_n$ is the space of the actuated joint vectors.

Let $g$ be the map relating the actuated joint vectors to the moving platform configurations :

$$\begin{aligned} g \quad &: OS_m \rightarrow JS_n \\ &X \;\; \rightarrow q = g(X) \end{aligned} \qquad (7)$$

It is assumed $m = n$, that is, only non-redundant manipulators will be studied in this paper. Finding the solutions $X$ to equation $q = g(X)$ means solving the direct kinematic problem. For our planar parallel manipulator, it has been shown that the direct kinematic model can admit six real solutions **[Gosselin 91]**.

Let $W$ be the reachable workspace, that is, the set of all positions and orientations reachable by the moving platform **[Kumar 92]**, **[Pennock 93]**. Let $Q$ be the reachable joint space that is, the set of all joint vectors reachable by actuated joints.

$$Q = \left\{ q \in JS_n , \forall i \leq n, q_{i\,min} \leq q_i \leq q_{i\,max} \right\}, Q \subset JS_n \qquad (8)$$

$$Q = g(W), W \subset OS_m \qquad (9)$$

**Definition 1:**

The aspects $WA_i$ are defined as the maximal sets such that :

- $WA_i \subset W$ ;
- $WA_i$ is connected ;
- $\forall X \in WA_i, \; Det(A) \neq 0$ .

In other words, the aspects are the maximal singularity-free connected regions *in the workspace.*

The aspects are computed as the connected components of the set obtained by removing the singularity surfaces from the workspace, which can be done easily with the octree model :

$$\cup WA_i = W - S \qquad (10)$$

**Application :**

For the planar manipulator studied, we get two aspects ($WA_1$ and $WA_2$), where $Det(A) > 0$ and $Det(A) < 0$, respectively. The singular surface of Figure 5 is the common boundary of the two aspects (Figure 6).

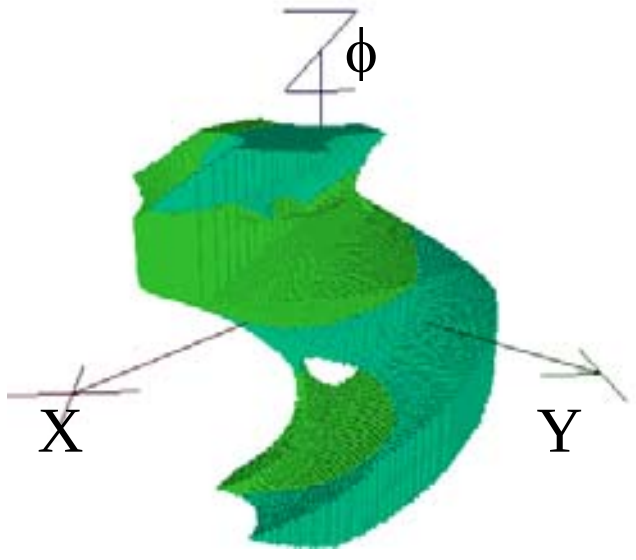

*Figure 6 : Octree model of the aspects*

### 3.2 NON-SINGULAR CONFIGURATION CHANGING TRAJECTORIES

In **[Innocenti 92]**, a non-singular configuration changing trajectory was found for the planar manipulator. However, it appears that this trajectory passes close to a singular configuration. We have been able to confirm that non-singular configuration changing trajectories do exist for this robot. For the following input joint values :

$$\rho_1 = 14.98 \quad \rho_2 = 15.38 \quad \rho_3 = 12.0 \qquad (11)$$

The solutions to direct kinematic problem are (Table 1) :

| | x | y | φ (rad) |
|---|---|---|---|
| 1 | -8.715 | 12.183 | -0.987 |
| 2 | -5.495 | -13.935 | -0.047 |
| 3 | -14.894 | 1.596 | 0.244 |
| 4 | -13.417 | -6.660 | 0.585 |
| 5 | 14.920 | -1.337 | 1.001 |
| 6 | 14.673 | -3.013 | 2.133 |

*Table 1 : Six direct kinematic solutions for the planar manipulator*

It can be verified that solutions -2-, -3-, -6- are in the same aspect ($WA_1$) (Figure 7).

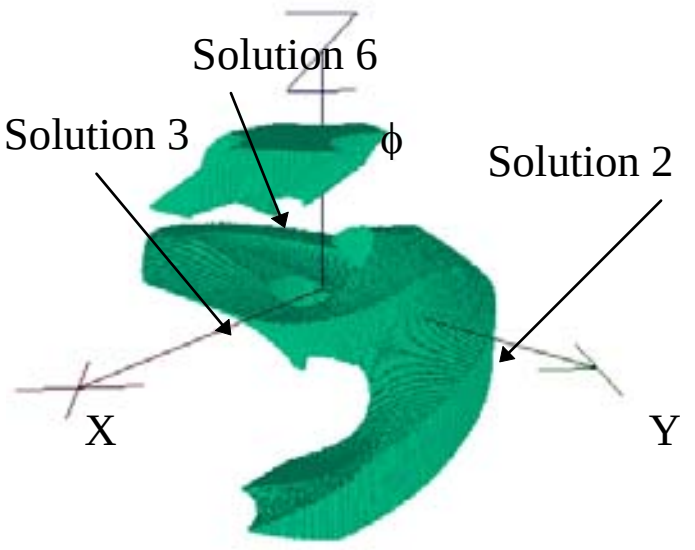

*Figure 7 : Three direct kinematic solutions in one single aspect*

This example clearly shows that the aspects are not the uniqueness domains. Additional surfaces have to be defined for separating the solutions.

# 4. CHARACTERISTIC SURFACES

## 4.1 DEFINITION

The characteristic surfaces were initially introduced in **[Wenger 92]** for serial manipulators.

This definition is restated here for the case of parallel manipulators.

**Definition 2 :**

Let *WA* an aspect in workspace *W*. The *characteristic surfaces* of the aspect *WA*, denoted $S_c(WA)$, are defined as the preimage in *WA* of the boundary $\overline{WA}$ that delimits *WA* (Figure 8) :

$$S_c(WA) = g^{-1}\left(g\left(\overline{WA}\right)\right) \cap WA \qquad \textbf{(12)}$$

where :

- $g$ is defined as in (15)
- $g^{-1}$ is a notation. Let $B \subset Q$ :
  $$g^{-1}(B) = \left\{ X \in W \,/\, g(X) \in B \right\}$$

The boundaries $\overline{WA}$ of *WA* are composed of :

- the type-2 singularities ;
- the type-1 singularities (limits of the actuated joints).

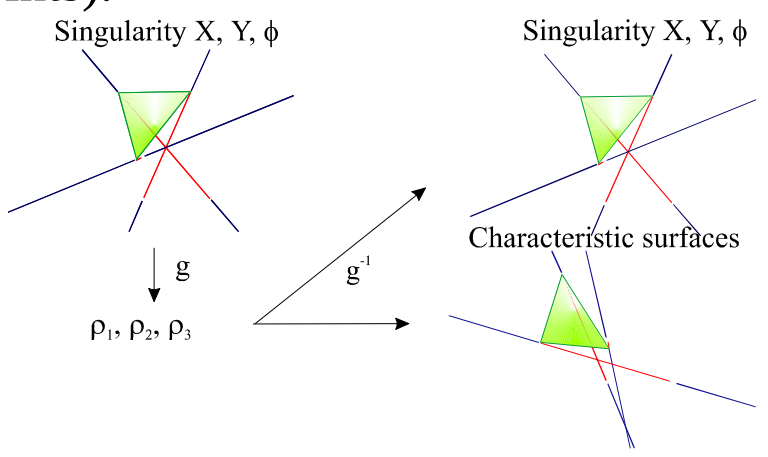

*Figure 8 : Definition of the characteristic surfaces*

## 4.2 CASE OF THE PLANAR 3-RPR MANIPULATOR

The characteristic surfaces are computed using definition (12). The singular surfaces are scanned and their preimages in the joint space are calculated using *g*. The resulting inverse singularities are mapped back into the workspace using the direct kinematic model. We get one characteristic surface for each aspect, denoted $S_{c1}$ and $S_{c2}$, respectively. Figure 9 depicts the singularity surface *S* along with the characteristic surface $S_{c1}$.

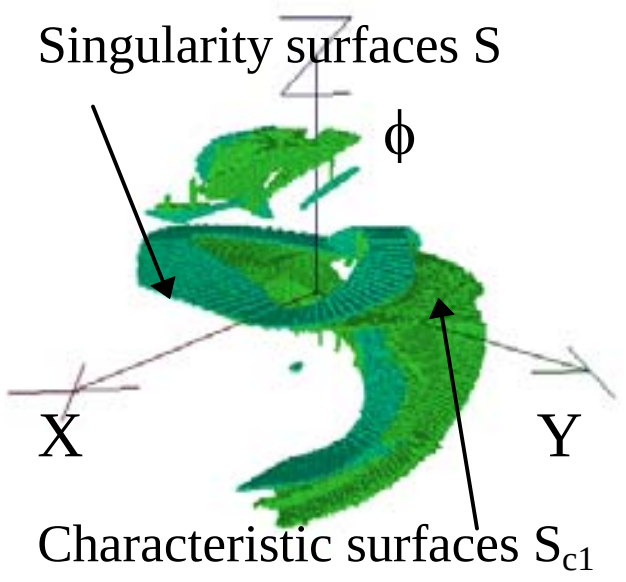

*Figure 9 : Octree model of the singularities and of the characteristic surfaces* $S_{c1}$

# 5. UNIQUENESS DOMAINS

## 5.1 BASIC COMPONENTS AND BASIC REGIONS

**Definition 3 :**

Let *WA* be an aspect. The *basic regions* of *WA*, denoted $\{WAb_i, i \in I\}$, are defined as the connected components of the set $WA \dot{-} S_C(WA)$ ($\dot{-}$ means the difference between sets). The *basic regions* induce a partition on *WA* :

$$WA = \left(\cup_{i\in I} WAb_i\right) \cup S_C(WA) \qquad \textbf{(13)}$$

**Definition 4 :**

Let $QA_{bi} = g(WAb_i)$, $QA_{bi}$ is a domain in the reachable joint space *Q* called *basic components*. Let *WA* an aspect and *QA* its image under *g*. The following relation holds :

$$QA = \left(\cup_{i\in I} QAb_i\right) \cup g\left(S_C(WA)\right) \qquad \textbf{(14)}$$

**Proposition 1 :**

The *basic components* of a given aspect are either coincident, or disjoint sets of *Q*.

**Theorem 1 :**

The restriction of *g* to any basic region is a bijection. In other words, there is only one direct solution in each basic region.

**Proof :**

We define the function *F* of $WA \times QA$ in *QA* (where $QA = g(WAb)$) such that :

$$\forall (X,q) \in WA \times QA,\ F(X,q) = g(X) - q \qquad \textbf{(15)}$$

*F* is a continuous and differentiable function on $WA \times QA$. Let $QAb_i$ be any basic component and let $WAb_i$ be such that $QAb_i = g(WAb_i)$.

Let $(X_0, q_0)$ be an arbitrary point in $WAb_i \times QAb_i$ such that $F(X_0, q_0) = 0$.

Since $(X_0, q_0) \in WAb_i \times QAb_i$, $Det\left(\frac{\partial F}{\partial X}(X_0, q_0)\right) \neq 0$.

Then, the implicit function theorem tells us that there exists a neighbourhood *U* of $X_0$ in $WAb_i$ and a neighbourhood *V* of $q_0$ in $QAb_i$ such that, for any *q* in $V$, equation $F(X, q)$ has one unique solution $X = f_1(q)$ in $U$.

Let $f_2$ be another solution to equation $F\left(f_2(q), q\right) = 0$ and such that $f_1(q_0) = f_2(q_0)$ . Since $q_0$ is the image of a configuration lying neither on the boundary nor on a singularity, $f_2$ is well defined at $q_0$, and $Det\left(\frac{\partial F}{\partial X}\left(f_2(q_0), q_0\right)\right) \neq 0$ .

If we prove that $f_1(q_0) = f_2(q_{0.})$ for any point $q$ in $QAb_i$, theorem 1 will be proved.

Let $C = \left\{q \in QAb_i \;/\; f_1(q) = f_2(q)\right\}$ : we have to prove that $C = QAb_i$. $C$ is a not empty since it contains points $q_0$. $C$ is a closed set in $QAb_i$ since $C$ is the set of all the roots of $f_2(q) - f_1(q) = 0$ .

Now $Det\left(\frac{\partial F}{\partial X}\left(f_2(q), q\right)\right) \neq 0$ for any point $q$ in $C$, since $QAb_i$ and thus $C$ does not contain points which are the image under $g$ of a singularity. Therefore, it stems from the implicit function theorem that $C$ has a neighbourhood of each of its points, and thus is also an open set in $QAb_i$. Since $QAb_i$ is connected, the only subsets of $QAb_i$ which are open an closed at the same time are $QAb_i$ and the empty set. Since $C$ is not empty, $C = QAb_i$ and the theorem is proved.

**Application :**

The basic regions are calculated as the connected components of the set obtained by removing the characteristic surfaces from the aspects :

$$\cup WA_{bi} = WA_i - S_{ci} \qquad \textbf{(16)}$$

We obtain thus 28 basic regions for the planar manipulator at hand (Figure 10).

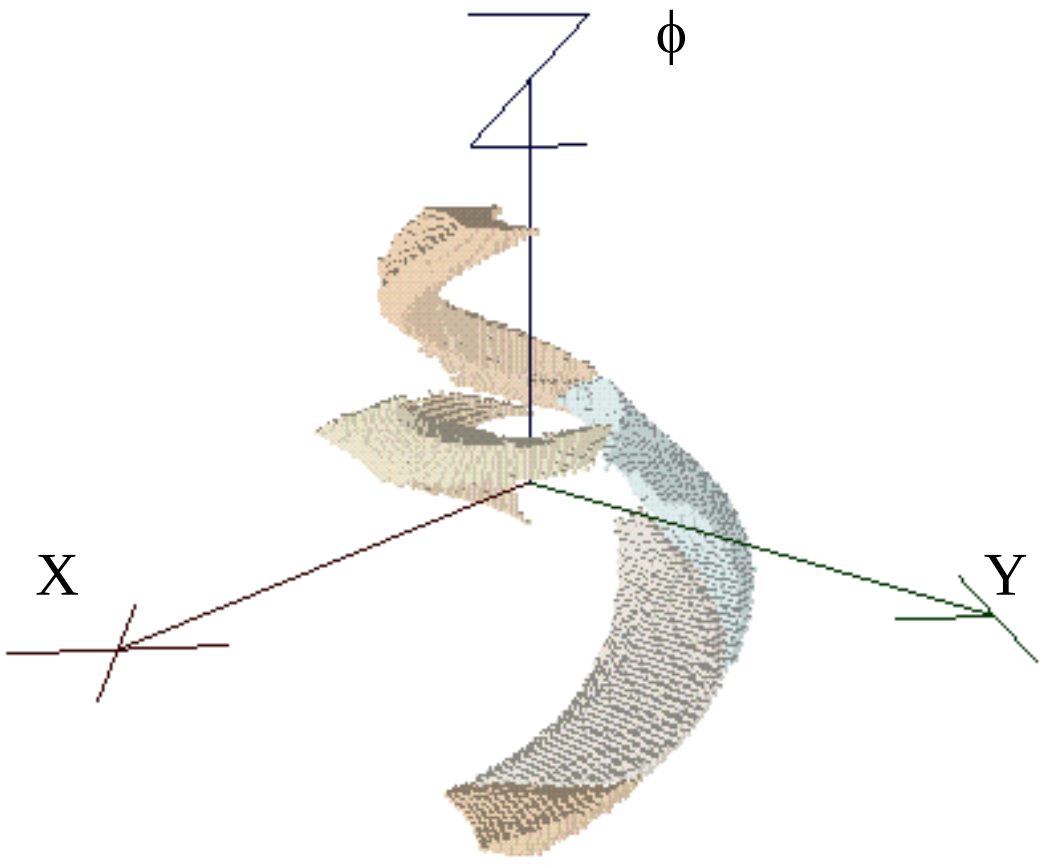

*Figure 10 : Octree model of the basic regions*

The basic components are computed as :

$$QA_i = g\left(WAb_i\right) \qquad \textbf{(17)}$$

We notice that, in accordance with proposition 1, the basic components are either coincident or disjoint. The coincident components yield domains with 2, 4 or 6 solutions for the direct kinematic problem : the upper one domains contain two coincident basic components, the middle five domains are composed of four coincident domains, and the last two domains contain six coincident basic components (Figure 11).

## 5.2 UNIQUENESS DOMAINS

Theorem 1 yields sufficient conditions for defining domains of the workspace (the basic regions) where there is one unique solution for the direct kinematic problem. However, the basic regions are not the maximal uniqueness domains. The following theorem 2 intends to define the larger uniqueness domains in the workspace.

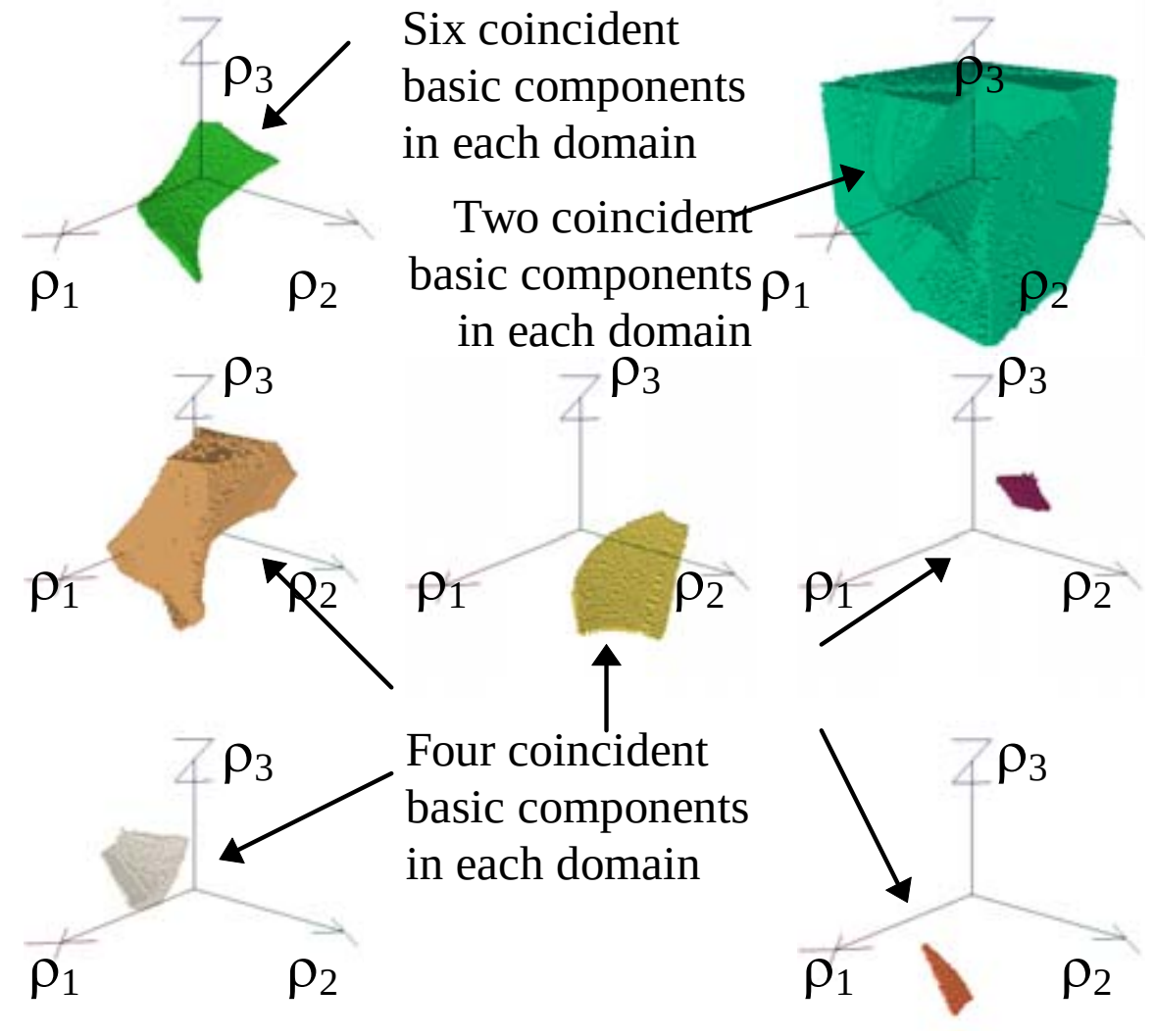

*Figure 11 : Octree model of the basic components*

**Theorem 2 :**

The uniqueness domains $Wu_k$ are the union of two sets : the set of adjacent basic regions $(\cup_{i \in I'} WAb_i)$ of the same aspect $WA$ whose respective preimages are disjoint basic components, and the set $S_c(I')$ of the characteristic surfaces which separate these basic components :

$$Wu_k = \left(\cup_{i \in I'} WAb_i\right) \cup Sc\left(I'\right) \qquad \textbf{(18)}$$

with $I' \subset I$ such as $\forall i_1, i_2 \in I'$, $g\left(WAb_{i_1}\right) \cap g\left(WAb_{i_2}\right) = \varnothing$ .

**Proof :**

Since the basic regions which define the $Wu_k$ are such that their preimages in the joint space do not overlap, there is still one unique solution in each $Wu_k$ .

**Application :**

To build the uniqueness domains, we have to consider the adjacent basic regions corresponding to disjoint, adjacent basic components.

Six uniqueness domains have been found for the planar manipulator (Figure 12), that is as many as the number of direct kinematic solutions. Note that in general, the number of uniqueness domains should be always more or equal to the maximal number of direct kinematic solutions. We notice that, in the joint space, a non-singular configuration changing trajectory has to go through a domain made of only two coincident basic components, that is, a domain where there are only two direct kinematic solutions (one in each aspect). In addition, it is worth noting that the domain with six coincident basic components map into six basic regions which are linked together by singular surfaces. This means that a non-singular configuration changing trajectory cannot be a straightforward motion between two such basic regions, but should pass through another intermediate basic region (Figure 13).

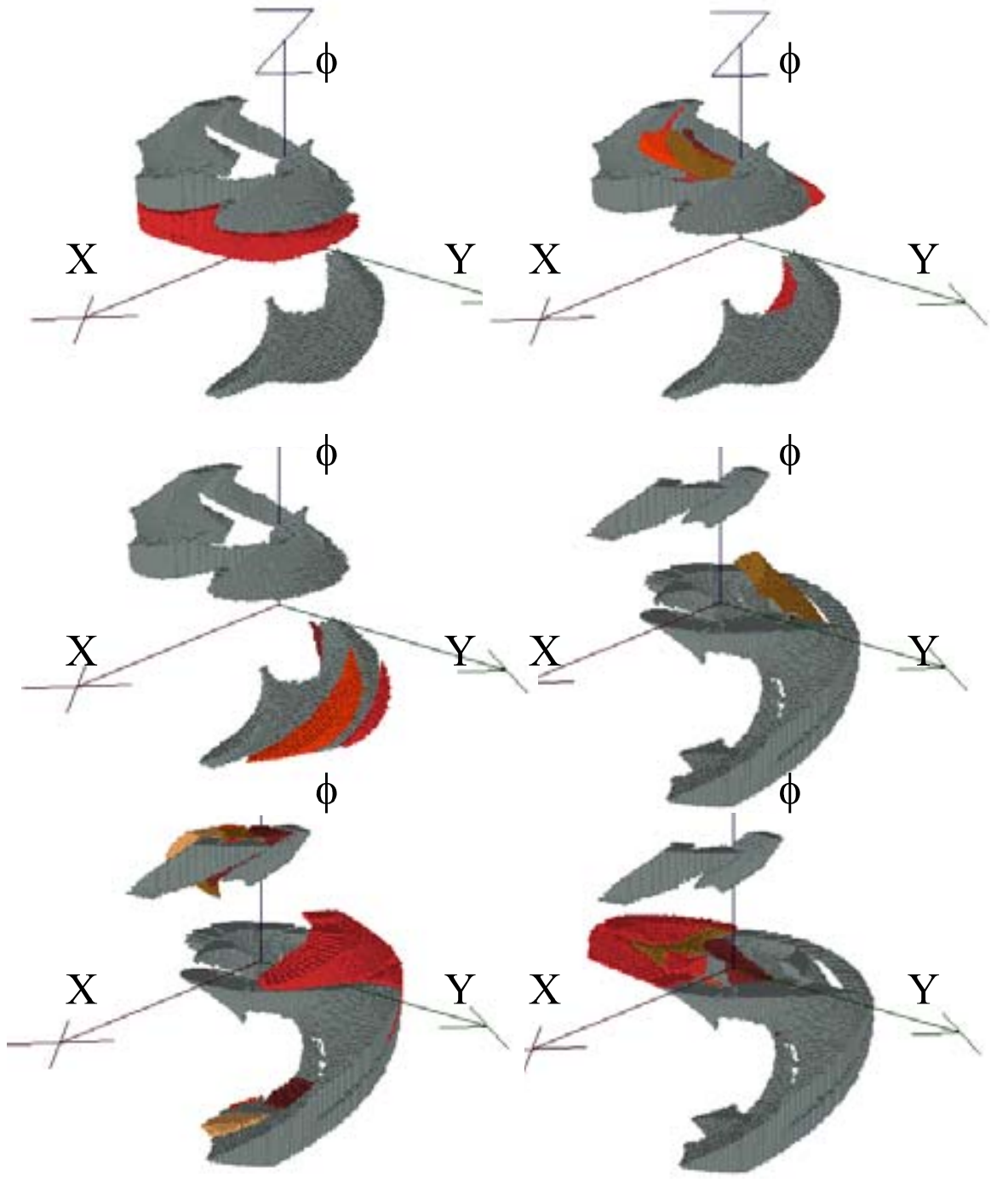

*Figure 12 : Octree model of the six uniqueness domains*

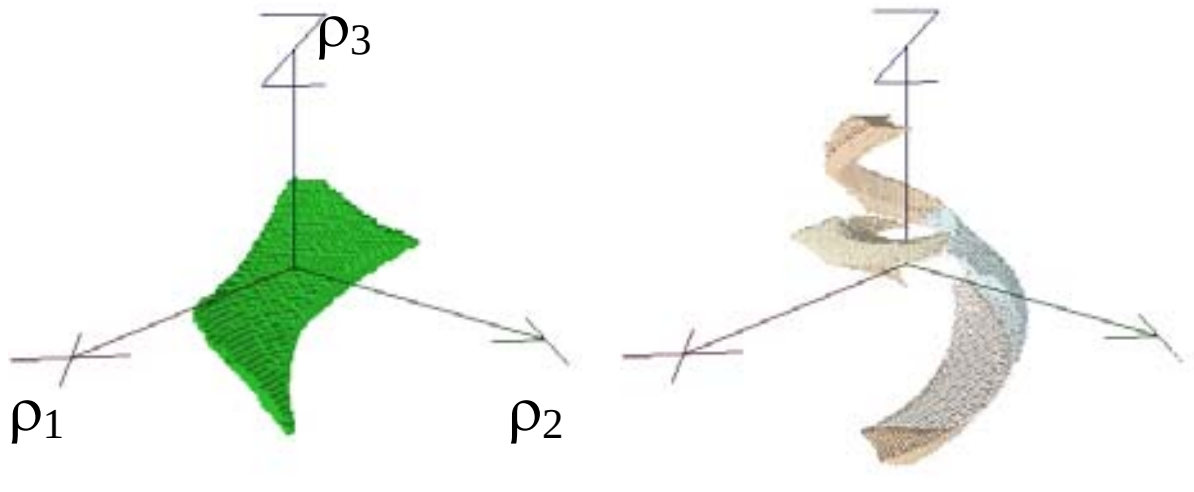

*Figure 13 : Basic regions and basic components with 6 solutions to the direct kinematic problem*

## 6. CONCLUSION

The problem of determining the uniqueness domains in the workspace of parallel manipulators has been studied in this paper. The aspects, originally introduced for serial manipulators, have been redefined here as the largest singularity-free regions in the workspace. The aspects were shown to be divided into distinct basic regions where there is only one solution to the direct kinematic problem. These regions are separated by the singular surfaces plus additional surfaces referred to as characteristic surfaces. Physically, the basic regions separate the different solutions to the direct kinematic problem : given a point in the joint space, the corresponding configurations of the moving platform are distributed in the different basic regions. The maximal uniqueness domains have been defined as the union of adjacent basic regions whose preimages in the joint space are not coincident. All results have been illustrated with a 3-DOF planar RPR-parallel manipulator. An octree model of space and specific enrichment techniques have been used for the construction of all sets. This work brings preliminary material to further investigations like trajectory planning **[Merlet 94]**, which is the subject of current research work from the authors.

## 7. BIBLIOGRAPHY

**[Borrel 86]** : Borrel Paul, «A study of manipulator inverse kinematic solutions with application to trajectory planning and workspace determination», Proc. IEEE Int. Conf. on Rob. And Aut., pp. 1180-1185, 1986.

**[Chablat 96]** : Chablat Damien, Wenger Philippe, « Domaines d'unicité pour les robots parallèles », Technical report, Laboratoire d'Automatique de Nantes, n°96.13, 1996.

**[Chételat 96]** : Chételat O., Myszkorowsky P., Longchamp R., « A Global Inverse Function Theorem for the Direct Kinematics », Proc. World Automation Congres-ISRAM'96, May 27-30, 1996, Montpellier, France.

**[El Omri 96]** : El Omri Jaouad, « Analyse Géométrique et Cinématique des Mécanismes de Type Manipulateur », Thèse, Nantes, 1996.

**[Faverjon 84]** : Faverjon B., « Obstacle avoidance using an octree in the configuration space of a manipulator », Proc. IEEE Int. Conf. on Rob. And Aut., pp. 504-510, 1984.

**[Garcia 86]** : Garcia G., Wenger P., Chedmail P., « Computing moveability areas of a robot among obstacles using octrees », Int. Conf. on Advanced Robotics (ICAR'89), June 1989, Columbus, Ohio, USA.

**[Gosselin 91]** : Gosselin Clément, Sefrioui Jaouad, Richard Marc J., « Solutions polynomiales au problème de la cinématique des manipulateurs parallèles plans à trois de gré de liberté », Mech. Mach. Theory, Vol. 27, pp. 107-119, 1992.

**[Innocenti 92]** : Innocenti C., Parenti-Castelli V., « Singularity-free evolution from one configuration to another in serial and fully-parallel manipulators », Robotics, Spatial Mechanisms and Mechanical Systems, ASME 1992.

**[Kumar 92]** : Kumar V., « Characterization of workspaces of parallel manipulators », ASME J. Mech. Design, Vol. 114, pp 368-375, 1992.

**[Khalil 96]** : Khalil W., Murareci D., « Kinematic analysis and singular configurations of a class of parallel robots », Mathematics and Computer in Simulation, pp. 377-390, 1996.

**[Meagher 81]** : D. Meagher, « Geometric Modelling using Octree Encoding », Technical Report IPL-TR-81-005, Image Processing Laboratory, Rensselaer Polytechnic Institute, Troy, 1981, New York 12181.

**[Merlet 90]** : Merlet J-P., « Les robots parallèles », HERMES, Paris, 1990.

**[Merlet 94]** : Merlet J-P., « Trajectory verification in the workspace of parallel manipulators », The Int. J. of Rob. Res, Vol 13, No 3, pp326-333, 1994.

**[Pennock 93]** : Pennock G.R., Kassner. D.J., « The workspace of a general geometry planar three-degree-of-freedom platform-type manipulator », ASME J. Mech. Design, Vol. 115, pp. 269-276, 1993.

**[Sefrioui 92]** : Sefrioui Jaouad, Gosselin Clément, « Singularity analysis and representation of planar parallel manipulators », Robots and autonomous Systems 10, pp. 209-224, 1992.

**[Wenger 92]** : Wenger Philippe, « A new general formalism for the kinematic analysis of all nonredundant manipulators », IEEE Robotics and Automation, pp. 442-447, 1992.